\documentclass{article}
\usepackage[utf8]{inputenc}
\usepackage{fullpage}
\usepackage{braket}

\usepackage{amsmath,amsfonts,amsthm,bm}

\def\eqref#1{equation~\ref{#1}}

\def\1{\bm{1}}

\DeclareMathAlphabet{\mathsfit}{\encodingdefault}{\sfdefault}{m}{sl}
\SetMathAlphabet{\mathsfit}{bold}{\encodingdefault}{\sfdefault}{bx}{n}

\newcommand{\R}{\mathbb{R}}

\usepackage{subfigure}

\usepackage{latexsym}
\usepackage{graphics}
\usepackage{amsmath}
\usepackage{xspace}
\usepackage{amssymb}
\usepackage{psfrag}
\usepackage{epsfig}
\usepackage{amsthm}
\usepackage{pst-all}

\usepackage{color}

\definecolor{Red}{rgb}{1,0,0}
\definecolor{Blue}{rgb}{0,0,1}
\definecolor{Olive}{rgb}{0.41,0.55,0.13}
\definecolor{Yarok}{rgb}{0,0.5,0}
\definecolor{Green}{rgb}{0,1,0}
\definecolor{MGreen}{rgb}{0,0.8,0}
\definecolor{DGreen}{rgb}{0,0.55,0}
\definecolor{Yellow}{rgb}{1,1,0}
\definecolor{Cyan}{rgb}{0,1,1}
\definecolor{Magenta}{rgb}{1,0,1}
\definecolor{Orange}{rgb}{1,.5,0}
\definecolor{Violet}{rgb}{.5,0,.5}
\definecolor{Purple}{rgb}{.75,0,.25}
\definecolor{Brown}{rgb}{.75,.5,.25}
\definecolor{Grey}{rgb}{.5,.5,.5}

\newcommand{\G}{\mathbb{G}}

\newcommand{\ignore}[1]{\relax}

\newtheorem{theorem}{Theorem}[section]

\newcommand{\ER}{Erd{\"o}s-R\'{e}nyi }

\definecolor{Red}{rgb}{1,0,0}
\definecolor{Blue}{rgb}{0,0,1}
\definecolor{Olive}{rgb}{0.41,0.55,0.13}
\definecolor{Green}{rgb}{0,1,0}
\definecolor{MGreen}{rgb}{0,0.8,0}
\definecolor{DGreen}{rgb}{0,0.55,0}
\definecolor{Yellow}{rgb}{1,1,0}
\definecolor{Cyan}{rgb}{0,1,1}
\definecolor{Magenta}{rgb}{1,0,1}
\definecolor{Orange}{rgb}{1,.5,0}
\definecolor{Violet}{rgb}{.5,0,.5}
\definecolor{Purple}{rgb}{.75,0,.25}
\definecolor{Brown}{rgb}{.75,.5,.25}
\definecolor{Grey}{rgb}{.5,.5,.5}
\definecolor{Pink}{rgb}{1,0,1}
\definecolor{DBrown}{rgb}{.5,.34,.16}
\definecolor{Black}{rgb}{0,0,0}

\usepackage{float}

\usepackage{amssymb}
\usepackage{mathtools}	
\usepackage{amsthm}
\usepackage{amsmath}

\usepackage{hyperref}
\usepackage{cleveref}

\usepackage{graphicx}
\usepackage{xcolor}

\theoremstyle{plain}

\title{Barriers for the performance of graph neural networks (GNN) in discrete random structures. 
A comment on~\cite{schuetz2022combinatorial},\cite{angelini2023modern},\cite{schuetz2023reply}}

\author{
{\sf David Gamarnik}    \thanks{Operations Research Center, Statistics and Data Science Center,  Sloan School of Management, MIT; e-mail: {\href{mailto:gamarnik@mit.edu}{\texttt{gamarnik@mit.edu}}}} \thanks{Funding from NSF Grant DMS-2015517 is gratefully acknowledged.}
}

\date{\today}

\begin{document}

\maketitle

\begin{abstract}

\end{abstract}


Recently graph neural network (GNN) based algorithms were proposed to solve a variety of combinatorial optimization problems, including Maximum Cut problem,
Maximum Independent Set problem and similar other problems~\cite{schuetz2022combinatorial},\cite{schuetz2022graph}. 
The algorithm was tested in particular on random instances of these problems, namely when
the underlying graph is generated according to some specified probability distribution. 
Earlier, a similar proposal using a somewhat different learning architecture was put forward to solve
another optimization problem, one of finding ground states of spin glass models~\cite{shen2022finding}. 

The publication~\cite{schuetz2022combinatorial} stirred a debate whether GNN based method was
adequately benchmarked against best prior methods. In particular, critical commentaries~\cite{angelini2023modern} and~\cite{boettcher2023inability} 
point out  that simple greedy algorithm
performs better than GNN in the setting of random graphs, and in fact stronger algorithmic performance can be reached with more sophisticated methods. 
A response from the authors~\cite{schuetz2023reply} pointed out that GNN performance can be improved further by tuning up the parameters better.

We do not intend to discuss the merits of arguments and counter-arguments 
in~\cite{schuetz2022combinatorial},\cite{angelini2023modern},\cite{boettcher2023inability},\cite{schuetz2023reply}. 
Rather in this note we  establish a fundamental limitation for running GNN
on random graphs considered in these references, for a broad range of choices of GNN architecture. Specifically, these barriers hold when the depth of GNN does 
not scale with graph size (we note that depth 2 was used in experiments in~\cite{schuetz2022combinatorial}), 
and importantly \emph{regardless} of any other parameters of GNN architecture,
including internal dimension and update functions. These limitations arise from the presence of the Overlap Gap Property (OGP) phase transition,
which is  a barrier for many  algorithms, both classical and 
quantum, including importantly local algorithms~\cite{gamarnik2021overlap},\cite{gamarnik2022disordered}. As we demonstrate in this paper, 
it is also a barrier to GNN due to its local structure.
We note that at the same time known algorithms ranging from   simple greedy algorithms to more sophisticated algorithms based on message passing, 
provide best results for these problems \emph{up to} the OGP phase transition. This leaves very little space for GNN to outperform the known algorithms, 
and based on this we side with the conclusions
made in~\cite{angelini2023modern} and~\cite{boettcher2023inability}.

\section{GNN for Combinatorial Optimization in Random Graphs}
A class of problems discussed in~\cite{schuetz2022combinatorial} and solved using GNN based methods 
falls into the domain of combinatorial optimization in random graphs. 
A graph $G$  is a collection of nodes $V$ and edges $E$, which is s subset of unordered pairs or, more generally, tuples (hyper-edges) 
of nodes. A generic combinatorial 
optimization problem is defined by introducing a cost function $C:\{0,1\}^V\to \R$ (also called Hamiltonian in physics jargon), 
which maps bit strings $\sigma\in \{0,1\}^V$ (aka ''decisions'') into real values $C(\sigma)$ (aka ''cost'' or ''energy''), and solving the problem
$\max_\sigma C(\sigma)$. An equivalent choice of $\sigma\in \{0,1\}^V$ will be adopted here often for convenience.
The presence of various kinds of combinatorial  constraints on decisions arising from the presence of edges and hyper-edges  
can be encoded into the cost function $C$.

A canonical example considered
in the aforementioned references is the Independent Set problem (which we abbreviate as IS)  
which is a problem of finding a largest in cardinality subset $I\subset V$
such that no two nodes are spanned by an edge. Namely $(i,j)\notin E$ for all $i,j\in I$. This corresponds to a special case of $C$
where $C(\sigma)=\left(\sum_{i\in V}\sigma_i\right) {\bf 1}\left( \sigma_i\sigma_j=0, \forall (i,j)\in E \right)$. Namely, $C$ is the number of ones
in the string $\sigma$ (indicating a inclusion into the independent set) multiplied by the indicator function for the event that $\sigma$ indeed encodes
a legitimate independent set. Another example discussed in the same collection of references is the graph Maximum Cut problem (which we abbreviate as MAXCUT). 
This is the problem of partitioning nodes of a graph into two sets which maximizes the number of crossed edges. Formally, this corresponds to the 
cost function $C:\{-1,1\}^V\to \R$ defined by $C(\sigma)=\sum_{(i,j)\in E} {\bf 1}\left(\sigma_i\sigma_j=-1\right)$. This model extends naturally
to hypergraphs as follows. A $K$-uniform hypergraph is a pair of a node set $V$ and a collection $E$ of hyperedges, where each hyperedge
is an unordered subset of $K$ nodes. Thus $2$-uniform hypergraph is just a graph. An extension of MAXCUT to hypergraphs is obtained
by considering the cost function $C(\sigma)=\sum_{(i_1,\ldots,i_K)\in E} {\bf 1}\left( \sigma(i_1)\sigma(i_2)\cdots \sigma(i_K)=-1\right)$.
The case $K=2$ again then reduces to the case of MAXCUT on graphs.

Our last example, arising  from the studies of spin glasses, corresponds fixing an order $p$ 
tensor $J=(J_{i_1,\ldots,i_p}, i_1,\ldots,i_p\in V)\in \R^{n\otimes p}$ and defining 
$C(\sigma)=\sum_{i_1,\ldots,i_p\in V} J_{i_1,\ldots,i_p}\sigma_{i_1}\sigma_{i_2}\cdots \sigma_{i_p}$ for each $\sigma\in \{-1,1\}^V$.  
The optimization problem is one of finding the value of $\max_\sigma C(\sigma)$.
Put it differently, this an unconstrained optimization problem on a complete weighted hyper-graph with hyper-edges $(i_1,\ldots,i_p), i_1,\ldots,i_p\in V$. 

In the random setting, either the cost function $C$ or the graph $G$ (or both) are generated randomly  according to some probability distribution.
The setting discussed in~\cite{schuetz2022combinatorial} is IS problem when the underlying graph a random $d$-regular graph on the set of
 $n$ nodes denoted for convenience
by $V=\{1,\ldots,n\}$.   $d$-regular means every node has exactly $d$ neighbors. The graph is generated uniformly at random from the space
of all $d$-regular graphs on $n$ nodes (see~\cite{janson2011random},\cite{frieze2015introduction} 
for some background regarding existence and constructions). 
The random graph constructed
this way will be denoted by $\G_d(n)$.
The setting of spin glasses corresponds to assuming that the entries of the tensor $J$ are generated randomly and independently
from some common distribution, such as the standard normal distribution.

Next we turn to a generic description of GNN algorithms. We follow the notations used in~\cite{schuetz2022combinatorial}. 
Given a graph $G=(V,E)$ the algorithm generates a sequence of node and time dependent
features $(h_{u,t}\in \R^{d_u}, u\in V, t\ge 0)$. Time is assumed to  evolve in discrete steps $t=0,1,2,\ldots$, and $d_u$
represents the dimension of the feature space for node $u$.
The feature vectors $h_{u,t}$ are generated as follows. The algorithm designer creates a node and time dependent   functions $(f_{u,t}, u\in V, t\ge 0)$
where each $f_{u,t}$ maps $\R^{d_u+\sum_{v\in \mathcal{N}(u)}}\to \R^{d_u}$. Here $\mathcal{N}(u)$ denotes the set of neighbors of $u$ 
( the set of nodes $v$ such that $(u,v)\in E$). The features are then  updated according to the rule
$h_{u,t+1}=f_{u,t}\left(h_{u,t}, \{h_{v,t}, v\in \mathcal{N}(u)\}\right)$. The update rules $f_{u,t}$ can be parametric or non-parametric (our conclusions
do not depend on that), and can be learned using various learning algorithms. The algorithm runs for a certain time $t=0,1,\ldots,R$, which
is also the depth of the underlying neural architecture. The obtained vector of features $(h_{u,R}, u\in V)$ is then projected to a desired solution of the problem. 
As we will see below, the actual details of how the update functions $f_{u,t}$ come about,
and, furthermore, regardless of the dimensions $d_u, u\in V$ that the algorithm designers opts to work with, the power of GNN algorithms is fundamentally
limited by the Overlap Gap Property, which we turn to next.

\section{Limits of GNN}
We begin with some  background on   problems introduced earlier: IS and MAXCUT in a setting of random  graphs,
and  ground states of spin glasses. Let $I_n^*$ denote (any) maximum size  independent set in $\G_d(n)$, which we recall
is a random $d$-regular graph,  and $|I_n^*|$ denote its size (cardinality). 
The following two facts were established
in~\cite{BayatiGamarnikTetali} and~\cite{frieze1992independence} respectively. 
For each $d$ there exists $\alpha_d$ such that $| I_n |/n$ converges to $\alpha_d$ with high probability as $n\to\infty$.
Furthermore, $\alpha_d=2(1+o_d(1))\log d/d$. Here $o_d(1)$ denotes a function which converges to zero as $d\to\infty$. 
Informally, we summarize this by saying that the size $| I^*_n|$ of a largest independent set in $\G_d(n)$ is approximately $2 (\log d/d)n$.

Next we turn to the discussion of algorithms for finding large independent sets in $\G_d(n)$.  It turns out that the best known 
algorithm for this problem is in fact the Greedy algorithm 
(the algorithms discussed in~\cite{angelini2023modern},\cite{boettcher2023inability}) which recovers a factor $1/2$-optimum independent set. More precisely, 
let $I_{\rm Greedy}$ be the independent set produced by the Greedy algorithm for $\G_d(n)$. 
Then $\lim_{d\to\infty} \alpha_d^{-1}\lim_n (| I_{\rm Greedy}|/n) = 1/2$ as $d\to\infty$,
Exercise 6.7.20 in~\cite{frieze2015introduction}.
No algorithm is known which beats Greedy by a factor non-vanishing in $d$. 

The theory based on the Overlap Gap Property (OGP) explains this phenomena rigorously. 
The OGP for this problem was established in~\cite{gamarnik2014limits} 
and it reads as follows: for every factor $1/2+1/(2\sqrt{2})<\theta<1$ there exists $0<\nu_1<\nu_2<1$
such that for every two independent sets $I_1,I_2$ which are $\theta$-optimal, namely $| I_1 |/n\ge \theta\alpha_d$, $| I_2 |/n\ge \theta\alpha_d$, 
it is the case that   either $ | I_1\cap I_2 |/n \le \nu_1$ or 
$ | I_1\cap I_2 |/n \ge \nu_2$, for all large enough $d$, with high probability as $n\to\infty$. 
Informally, every two sufficiently large  independent sets (namely those  which are multiplicative factor $\theta$-close to optimality) are either ''close''
to each other (overlap in at least $\nu_2 n$ many nodes) or ''far'' from each other (overlap in at most $\nu_1 n$ many nodes).  Namely, solutions to the IS
optimization problem with sufficiently large optimization values exhibit a gap in the overlaps (hence the name of the property). 

It turns out that OGP is a barrier to a broad class of algorithms, in particular algorithms which are local in an appropriately defined sense.
This was established in the same paper~\cite{gamarnik2014limits}.
We introduce the notion of locality only informally. The formal definition involves the concept of Factors of IID for which we refer the reader
to~\cite{gamarnik2014limits}.  An algorithm, which maps graphs $G$ to an independent set in $G$ is called $R$-local if  for every node $u$ of the graph $G$,
the algorithmic decision as to whether to make this node a part of the constructed independent set or not, is based entirely on the size $R$ neighborhood of this 
node $u$. In particular, we see that  the GNN algorithm is $R$-local provided that the number of iterations $t$ of GNN is at most $R$. Importantly this holds
\emph{regardless} of the complexity of the feature dimensions $d_u$ and the choice of update functions $f_{u,t}$.
 We recall that the GNN algorithm reported in~\cite{schuetz2022combinatorial}
 was based on 2 iterations and as such it is 2-local.

A main theorem proved in~\cite{gamarnik2014limits} states that OGP is a barrier for all $R$-local algorithms, as long as $R$ is any constant not growing with the size
of the graph. Specifically, for any $R$, consider \emph{any} algorithm $\mathcal{A}$ which is $R$-local. Then the independent set produced
by $\mathcal{A}$ is at most $(1/2+1/(2\sqrt{2}))\alpha_d$ for large enough $d$ with high probability as $n\to\infty$. Using a more sophisticated notion 
of multi-overlaps the result was improved in~\cite{rahman2017local} to factor $1/2$ of optimality for the same class of all local algorithms. 
Importantly, as we  recall, $1/2$ is the threshold achievable by the Greedy algorithm. 
The result was recently extended to the class of algorithms based on low-degree polynomials 
and small depth Boolean circuits in~\cite{gamarnik2020lowFOCS},\cite{wein2022optimal}. 
It is
conjectured that beating the $1/2$ threshold is not possible within the class of polynomial time algorithms (but showing this will amount to proving $P\ne NP$). 

As a consequence of the discussion above we obtain an important conclusion regarding the power of GNN for solving the IS problem in $\G_d(n)$.

\begin{theorem}
Consider any architecture of the GNN algorithm with any choice of dimensions $(d_v, v\in \{1,2,\ldots,n\})$,  any choice of 
feature functions $h_{u,t}$, and any choice of update functions $f_{u,t}$. Suppose the GNN algorithm
 iterates for $R$ steps and produces an independent set $I_{\rm GNN}$ in the random
regular graph $\G_d(n)$. Then the size of $I_{\rm GNN}$ is at most  half-optimum asymptotically in $d$, for any  value of $R$.
\end{theorem}
We stress here that the depth parameter $R$ can be arbitrarily large and, in particular,  
may depend on the average degree $d$, provided it does not depend on the size
$n$ of the graph. 
We recall that $R=2$ in the implementation reported in~\cite{schuetz2022combinatorial}.
Since the Greedy algorithm already achieves $1/2$ optimality, as we have remarked earlier, this result leaves very little space for GNN to outperform the known (Greedy)
algorithm for the IS defined on random regular graphs. We note that while the results above are stated in the asymptotic sense of increasing degrees $d$,
the fact is that OGP is a barrier to local algorithm as soon as OGP holds. For example if it is the case   that say $\G_{10}(n)$ graph
exhibits the OGP above some approximation factor $\rho$ to optimality,  this would imply   that GNN  cannot beat the 
$\rho$-factor approximation for the IS problem in the random graph $\G_{10}(n)$ for any graph independent depth (number of rounds) $R$. 
The obstruction to this is proving the OGP for small values of $d$,
which is more  challenging mathematically.

Next we turn to the MAXCUT problem on random graphs and random hypergraphs. The situation here is rather similar, but better developed in the context
of random \ER graphs and hypergraphs, as opposed random regular graphs, and thus, this is the class of random graphs we now turn to. A random \ER graph
with average degree $d$ denoted by $\G(n,d)$ is obtained by connecting every pair of nodes $i,j$ among $n$ nodes with probability $d/n$, independently across
all unordered pairs $i\ne j$. A random $K$-uniform hypergraph is obtained similarly by creating a hyperedge from a collection of nodes $i_1,\ldots,i_K$ with
probability $d/{n-1 \choose K-1}$. We denote this graph by $\G(n,d;K)$ It is easy to see that the average degree in both $\G(n,d)$ and $\G(n,d;K)$ is $d+o(1)$. 
It was known for a while that the optimum values of MAXCUT in $\G(n,d;K)$ are of the form $n(d/(2K)+\gamma_K^*\sqrt{d}+o(\sqrt{d}))$ 
as $n\to\infty$,~\cite{Coppersmith2004}, for some constant $\gamma_K$.  Namely, the optimum value is known up to the order $n\sqrt{d}$. The constant
$\gamma_K^*$ was computed in~\cite{dembo2017extremal} and
\cite{sen2018optimization} first for the case $K=2$ and then extended to general $K$ in~\cite{chen2019suboptimality}. 
As it turns out, this constant is the value of the ground state of a $K$-spin model, known 
since the work of Parisi~\cite{parisi1980sequence}, Talagrand~\cite{talagrand2006parisi} and Panchenko~\cite{panchenko2013sherrington}. 

Interestingly, as far as the algorithms are concerned there is a fundamental difference between the case $K=2$ (aka graphs) versus $K\ge 3$. 
Specifically, algorithms achieving the asymptotically optimal value 
$n(d/(2K)+\gamma_K^*\sqrt{d}+o(\sqrt{d}))$ are known based on Approximate Message Passing (AMP)
schemes~\cite{alaoui2021local}. Furthermore, conjecturally, the OGP  does not hold for this problem. 
However, when $K\ge 4$ and is even, OGP provably does hold and again 
presents a barrier to all local algorithms, as was established in~\cite{chen2019suboptimality}. 
Furthermore, a sophisticated version of the multi-OGP called Branching-OGP 
was computed~\cite{huang2022tight}, the threshold for which, denoted by $\gamma_{\rm B-OGP,K}$
matches the best known algorithms, which is again the AMP type. 
The formal statement of the OGP is very similar to the one for the IS and we skip it. 
As an implication we obtain our second conclusion.

\begin{theorem}
Consider any architecture of the GNN algorithm which  produces a partition  $\sigma_{\rm GNN}\in \{\pm 1\}^n$ in the random
hyper graph $\G(n,d;K)$. Suppose $K\ge 4$ and is even. 
For sufficiently large degree values $d$ the size of the cut associated with this solution is at most
$n(d/(2K)+\gamma_{\rm B-OGP,K}\sqrt{d}+o(\sqrt{d}))$ with high probability, for any choice of $R$.  This is suboptimal since $\gamma_{\rm B-OGP,K}<\gamma_K^*$.
\end{theorem}
As the threshold $\gamma_{\rm B-OGP,K}$ is achievable by the AMP algorithm, again this leaves very little space for GNN to outperform the
best known (namely AMP) algorithm for this problem.

The story for the problem of finding near ground states in spin glasses is very similar and is skipped. 
We refer the reader to surveys~\cite{gamarnik2021overlap} and~\cite{gamarnik2022disordered} for details.
In fact many of the results above described for  the MAXCUT problem,
were obtained first by  deriving them for the spin glasses model, and then transferred to the case of random graphs $\G(n,d/n;K)$ using 
 an interpolation technique.

\section{Discussion}
In this paper we have presented a  barriers faced by GNN based algorithms in solving combinatorial optimization problems in random graphs and
random structures. These barriers stem from the complex solution space geometry property in the form of the Overlap Gap Property (OGP), a known barrier to broad
classes of algorithms, local algorithms in particular. As GNN falls within the framework of local algorithms, OGP is a barrier to GNN as well. Since algorithms
are known which achieve all the optimality values below the OGP phase transition threshold, this leaves very little room for GNN to outperform the known algorithms. 

Some further investigation can be done however to obtain a more refined picture. Most of the OGP results are obtained in the doubly-asymptotic regime where not 
only the graph size diverges but also the degree (and the degree type parameters) diverge. While it is possible to prove  OGP for a fixed and sufficiently large values
of the degree, 
the values arising from these computations tend to be quite large. Instead, it would be nice to see whether OGP takes place say in random 
regular graphs $\G_d(n)$ for a small degree value such as say $d=5$. We need sharper mathematical techniques for this. Knowing this might provide us
with a place where non-trivial algorithms going beyond the simple Greedy algorithms might provide some value. It is known 
(as already observed in~\cite{angelini2023modern}) 
that more clever version of the Greedy algorithm known as the Degree Greedy algorithm provably outperform the naive Greedy algorithms
for fixed values of $d$. It is possible thus that a more sophisticated
version of the GNN can perhaps achieve performance values even stronger than the ones obtained by the Degree Greedy algorithm. 
Whether this is possible is yet to be seen, but in case this is indeed verified rigorously, it would provide
a more compelling argument in favor of GNN than the one currently presented in~\cite{schuetz2022combinatorial}.

\bibliographystyle{alpha}
\bibliography{bibliography-04.2023}

\end{document}